\documentclass[conference]{IEEEtran}
\IEEEoverridecommandlockouts
\usepackage{cite}
\usepackage{amsmath,amssymb,amsfonts}
\usepackage{algorithmic}
\usepackage{graphicx}
\usepackage{textcomp}
\usepackage{xcolor}

\usepackage{witharrows}

\usepackage[flushleft]{threeparttable}
\usepackage{tablefootnote}
\usepackage{array}
\newcolumntype{P}[1]{>{\centering\arraybackslash}p{#1}}
\usepackage{multicol}
\usepackage{multirow}
\usepackage{tabulary}
\usepackage{colortbl}
\definecolor{mygray}{gray}{0.90}
\usepackage{makecell}
\usepackage{booktabs}
\usepackage{subcaption}

\usepackage{stmaryrd}
\usepackage{mathabx}

\usepackage{url,hyperref,microtype}
\usepackage[switch,columnwise]{lineno}
\def\BibTeX{{\rm B\kern-.05em{\sc i\kern-.025em b}\kern-.08em
    T\kern-.1667em\lower.7ex\hbox{E}\kern-.125emX}}

\usepackage{fancyhdr}
\begin{document}

\title{Synthetic Thermal and RGB Videos for Automatic Pain Assessment utilizing a Vision-MLP Architecture\\
\thanks{$\dagger$: Corresponding Author, $\ddagger$: Affiliated Researcher at the Computational
Biomedicine Laboratory of the Foundation for Research and Technology
(FORTH), Heraklion, Greece}
}

\author{\IEEEauthorblockN{Stefanos Gkikas$^\dagger$}
\IEEEauthorblockA{\textit{Department of Electrical \& Computer Engineering} \\
\textit{Hellenic Mediterranean University}\\
Heraklion, Greece\\
gkikas@ics.forth.gr}
\and
\IEEEauthorblockN{Manolis Tsiknakis$^\ddagger$}
\IEEEauthorblockA{\textit{Department of Electrical \& Computer Engineering} \\
\textit{Hellenic Mediterranean University}\\
Heraklion, Greece \\
tsiknaki@ics.forth.gr}

}

\maketitle

\begin{abstract}
Pain assessment is essential in developing optimal pain management protocols to alleviate suffering and prevent functional decline in patients. Consequently, reliable and accurate automatic pain assessment systems are essential for continuous and effective patient monitoring.
This study presents synthetic thermal videos generated by \textit{Generative Adversarial Networks} integrated into the pain recognition pipeline and evaluates their efficacy.
A framework consisting of a \textit{Vision-MLP} and a \textit{Transformer}-based module is utilized, employing RGB and synthetic thermal videos in unimodal and multimodal settings.
Experiments conducted on facial videos from the \textit{BioVid} database demonstrate the effectiveness of synthetic thermal videos and underline the potential advantages of it.

\end{abstract}


\section{Introduction}
Pain, as defined by the International Association for the Study of Pain (IASP), is \textit{\textquotedblleft an unpleasant sensory and emotional experience associated with actual or potential tissue damage, or described in terms of such damage\textquotedblright} \cite{iasp_2020}. Additionally, Williams and Craig \cite{williams_craig_2016} state that pain encompasses emotional, cognitive, and social dimensions beyond physical aspects. 
Biologically, pain is a distasteful sensation that originates in the peripheral nervous system. It serves a vital function by activating sensory neurons, alerting the body to potential injury, and playing a critical role in recognizing and reacting to hazards. \cite{khalid_tubbs_2017}. 
Pain is a significant concern impacting individuals and 
social structures. Daily, people across all age groups suffer from pain resulting from accidents, illnesses, or as a part of medical treatment, making it the most common cause for seeking medical consultation. Both acute and chronic pain present clinical, economic, and social challenges. 
In addition to its immediate impact on a patient's life, pain is related to several adverse outcomes, including opioid consumption, drug misuse, addiction, deteriorating social relationships, and mental health issues \cite{dinakar_stillman_2016}.
Effective pain assessment is crucial for early diagnosis, disease advancement monitoring, and treatment effectiveness, particularly in chronic pain management \cite{gkikas_tsiknakis_slr_2023}. Consequently, pain is referred to as \textit{\textquotedblleft the fifth vital sign\textquotedblright} \space in nursing literature \cite{joel_lucille_1999}. Objective measurement of pain is vital for providing appropriate care, especially for groups who cannot express their pain, such as infants, young children, people with mental health issues, and seniors. Numerous methods are utilized for pain assessment, including self-reporting, which remains the gold standard for determining pain presence and intensity through rating scales and questionnaires. 
Additionally, behavioral indicators such as facial expressions, vocalizations, and body movements are essential for evaluating pain. \cite{rojas_brown_2023}. Furthermore, physiological indicators such as electrocardiography, electromyography, skin conductance, and breathing rates provide significant insights into pain's physical effects \cite{gkikas_tsiknakis_slr_2023}.
Although pain evaluation holds significant importance, it poses a substantial challenge to medical practitioners \cite{aqajari_cao_2021}, particularly with patients who cannot communicate verbally. In senior patients, the situation becomes even more complex due to their decreased expressiveness or reluctance to share their pain experiences \cite{yong_gibson_2001}.
Furthermore, extensive studies \cite{bartley_fillingim_2013,gkikas_chatzaki_2022, gkikas_chatzaki_2023} reveal distinct disparities in pain expression among various genders and age categories, underlining the intricacy of the pain assessment process.

In recent years, there has been a growing tendency in the field of affective computing research to incorporate thermal imaging techniques \cite{qudah_2021}. The interest was sparked following findings in the literature that stress and cognitive load significantly impact skin temperature \cite{ioannou_2014}. This is due to the role of the autonomic nervous system (ANS) in controlling physiological signals like heart rate, respiration rate, blood perfusion, and body temperature, which are indicative of human emotions and affects \cite{qudah_2021}. Additionally, muscle contractions influence facial temperature by transferring heat to the facial skin \cite{jarlier_2011}.
Therefore, thermal imaging is a promising method for measuring transient facial temperatures \cite{merla_2004}.
The authors in \cite{youssef_2023} examined thermal imaging and facial action units to assess emotions, including frustration, boredom, and enjoyment. The multimodal approach demonstrated the highest accuracy. 
Thermal imaging has been explored in a relatively small number of studies within the field of pain research. In \cite{erel_ozkan_2017}, the authors observed that facial temperature rises following a painful stimulus, indicating that thermal cameras could serve as valuable tools for pain monitoring. In \cite{haque_2018}, a pain dataset comprising RGB, thermal, and depth videos was introduced. The findings demonstrated that the RGB modality marginally surpassed the others in performance while integrating all modalities led to superior results.

This study introduces the generation of synthetic thermal videos through generative adversarial networks (GANs), which are used in unimodal and multimodal settings combined with the RGB video modality. The foundation of the automatic pain assessment pipeline is a framework that integrates a Vision Multilayer Perceptron (MLP) model with a transformer-based module.
The primary contributions of our research include: (1) generating synthetic thermal videos to supplement pain assessment as an additional vision modality, (2) evaluating the effectiveness of RGB and synthetic thermal videos as independent modalities, (3) exploring the effectiveness of the thermal-related information, and (4) analyzing the performance and application of the newly introduced Vision-MLP architectures.

\section{Related Work}
\label{related_work}
Recent developments have seen a range of innovative methods to assess pain levels from video data. 
Werner \textit{et al.} \cite{werner_2016} focused on domain-specific features, using facial action markers with a deep random forest (RF) classifier, and proposed a 3D distance computation method among facial points while in \cite{werner_hamadi_walter_2017}, an optical flow method was introduced to track facial points and capture expression changes across frames. 
The dynamic aspects of pain were addressed by developing long short-term memory networks combined with sparse coding (SLSTM) \cite{zhi_wan_2019}. 
Tavakolian \textit{et al.} \cite{tavakolian_hadid_2019} utilized 3D convolutional neural networks (CNNs) with varied temporal depths to analyze short-, mid-, and long-term facial expressions. 
In \cite{thiam_kestler_schenker_2020}, the authors leverage the temporal aspect of videos by encoding frames into motion history and optical flow images, which were then analyzed using a combination of CNN and bidirectional LSTM (biLSTM).
Another method encoded videos into single RGB images through statistical spatiotemporal distillation (SSD) and trained a Siamese network in a self-supervised manner \cite{tavakolian_bordallo_liu_2020}. 
In \cite{huang_xia_li_2019} the authors implemented a multi-stream CNN for feature extraction from different facial regions, applying learned weights to emphasize the significance of each region's features in expressing pain. 
Further research \cite{huang_xia_2020} identified that specific frames more clearly displayed pain expressions and developed a framework using CNNs, gated recurrent units (GRUs), and attention saliency maps, assigning weights to each frame's influence on overall pain intensity. 
A novel approach by Huang \textit{et al.} \cite{huang_dong_2022} was introduced by extracting simulated heart rate data from video content utilizing a 3D CNN, demonstrating strong results in binary and multiclass classification scenarios.
Finally, in the studies \cite{gkikas_tsiknakis_embc, gkikas_tachos_2024}, transformer-based frameworks were proposed, yielding promising results with high efficiency.

\section{Methodology}
This section describes the process of generating synthetic thermal videos, the architecture of the proposed automatic pain assessment framework, the developed augmentation techniques, the pre-processing methods, and the pre-training strategy for the modules.

\subsection{Synthetic Thermal Videos}
\begin{figure}[h]
\begin{center}
\includegraphics[scale=0.16]{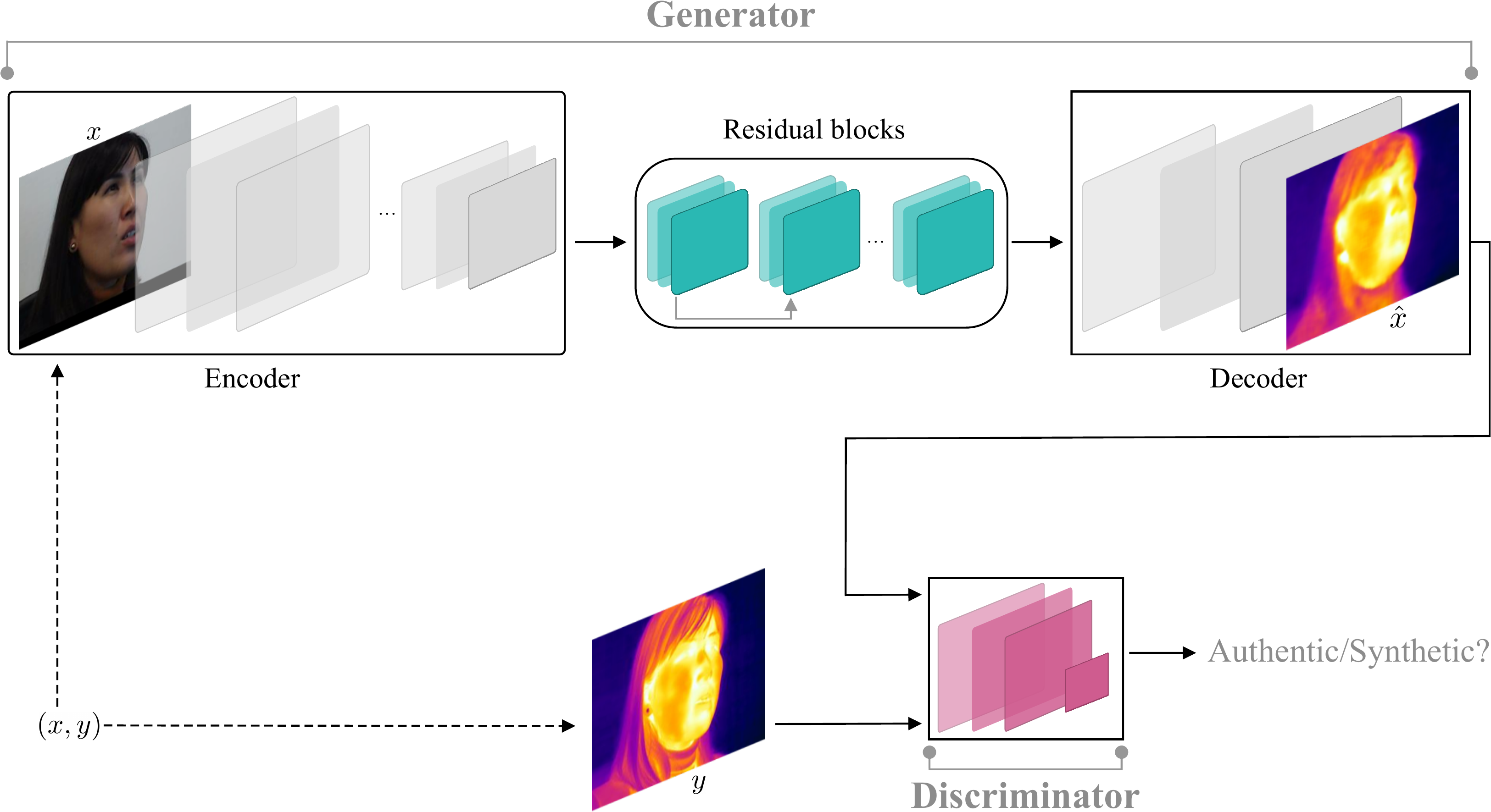}
\end{center}
\caption{Overview of the pipeline for generating synthetic thermal images, including the Generator 
$G$ (Encoder, intermediate ResNet, Decoder), and the Discriminator $D$.}
\label{gans}
\end{figure}

An image-to-image translation (I2I) approach has been developed for generating synthetic thermal videos. I2I generative models aim to map distinct image domains by learning the intrinsic data distributions of both domains. In this case, the source domain consists of RGB images, while thermal images represent the target domain.
In this study, conditional generative adversarial networks (cGANs) \cite{mirza_2014} were developed and trained in supervised settings with aligned image pairs. Fig. \ref{gans} illustrates a high-level overview of the proposed method. The generator $G$ generates realistic-looking images, while discriminator $D$ aims to distinguish authentic images from synthetic ones via the following minimax game:
\begin{equation}
\min_{G} \max_{D} \mathcal{L}_{\text{cGAN}}(G, D),
\end{equation}
where the objective function $\mathcal{L}_{\text{cGAN}}(G, D)$ can be expressed as: 
\begin{equation}
\mathbb{E}_{x, y}[\log D(x, y)] + \mathbb{E}_{x, z}[\log(1 - D(x, G(x, z)))],
\end{equation}
where $x$ represents the real data, $y$ signifies the target data, and $z$ denotes the random noise vector. The $G$ aims to minimize the objective function, while the $D$ functions adversarially, trying to maximize it. 
Furthermore, we included the Wasserstein gradient penalty (WGAN-GP) \cite{wgans} to increase the training stability. The final objective is described as:
\begin{equation}
\mathcal{L}_{\text{cGAN}}(G, D) + \lambda \mathbb{E}_{\hat{x}, y}[(\|\nabla_{\hat{x}} D(\hat{x}, y)\|_2 - 1)^2],
\end{equation}
where $\lambda$ denotes the penalty coefficient.
Regarding the architecture, in the proposed methodology, inspired by \cite{wang_liu_2018}, the $G$ is structured into $3$ distinct modules: an encoder, which comprises $2$ convolutional layers downsampling the input; an intermediate ResNet module, consisting of $9$ residual blocks, each consisting with $2$ convolutional layers; and a decoder, upsampling the feature maps to the final resolution (\textit{i.e.,} $256\times256$) for the synthetic sample. The $D$ founded on \cite{isola_2017} is a pixel-level PatchGAN discriminator using $1\times1$ kernels consisting of $2$ convolutional layers.

\begin{figure*}[h]
\begin{center}
\includegraphics[scale=0.8]{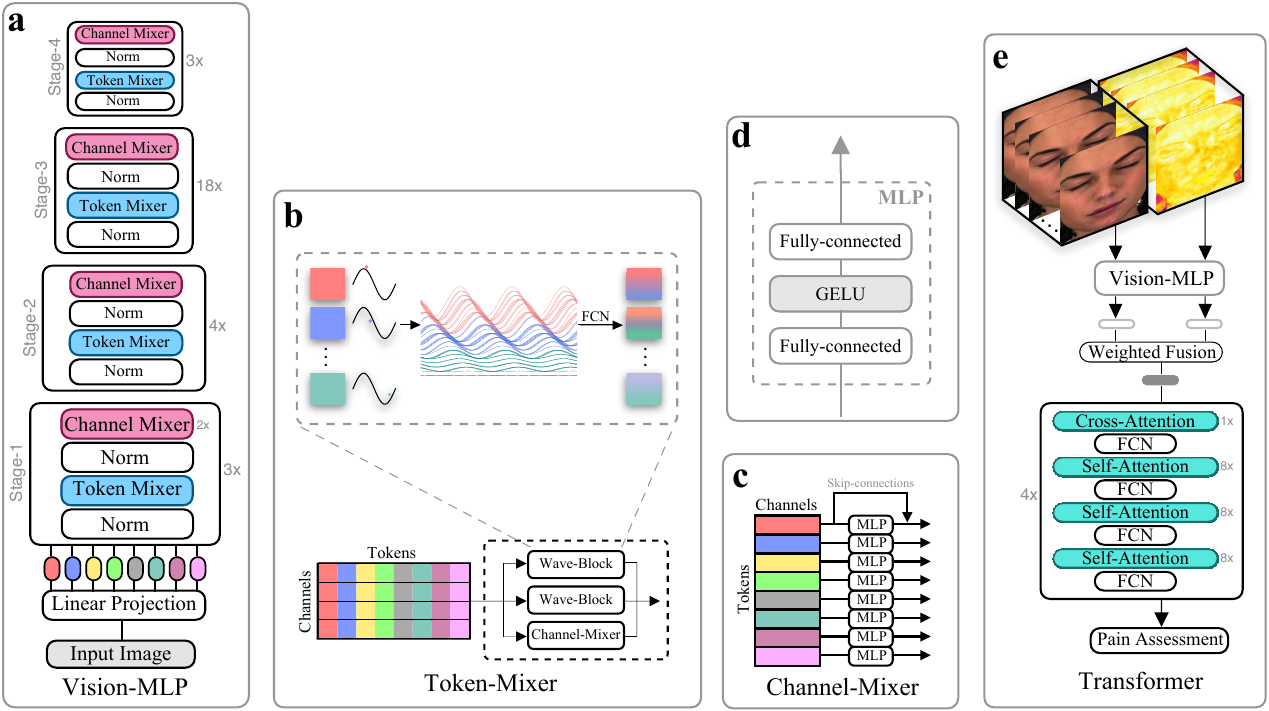}
\end{center}
\caption{Schematic overview of the proposed framework for automatic pain assessment, detailing its modules and their primary components: \textbf{(a)} The \textit{Vision-MLP} module, responsible for extracting embeddings from individual frames. \textbf{(b)} The \textit{Token-Mixer}, another major sub-module of \textit{Vision-MLP}, creates the wave representation for the tokens. \textbf{(c)} The \textit{Channel-Mixer}, a key sub-module within \textit{Vision-MLP}. \textbf{(c)} The MLP, an integral part of the \textit{Channel-Mixer}. \textbf{(e)} The fusion process combining RGB and synthetic thermal embeddings, followed by the \textit{Transformer} module, which performs the final pain assessment.}
\label{framework}
\end{figure*}

\subsection{Framework Architecture}
The proposed framework comprises two main modules: a Vision-MLP model that functions as a spatial embedding extractor for individual video frames and a transformer-based model that serves as a temporal module, utilizing the embedded representations of the videos for temporal analysis and final pain assessment. Fig. \ref{framework} illustrates the modules and their primary building blocks.

\subsubsection{Vision-MLP}
MLP-like models represent a newly introduced type of vision models, serving as alternatives to traditional Convolutional Neural Networks (CNNs) and Vision Transformers (ViT). They are characterized by simple architectures consisting of fully connected layers coupled with activation functions. They embody lesser inductive bias and are based on basic matrix multiplication routines. Our approach is founded on the principles of \cite{mlp_mixer} introducing the Vision-MLP and \cite{wave_vision_mlp} incorporating a wave representation for the patches (also referred to as tokens). 
Each video frame is initially divided into $n$ non-overlapping tokens 
$\mathcal{F}_{m}  = [f_{m,1}, f_{m,2}, \ldots, f_{m,n}]\in\mathbb{R}^{n\times p\times p\times 3 }$, where $p$ specifies the resolution of each token, \textit{i.e.,} $16\times16$ pixels, and $3$ represents the number of color channels. 
Each token is then linearly projected into a dimension $d=768$ before being fed into the \textit{Vision-MLP} (refer to \hyperref[framework]{Fig. 2a}).
The first main sub-module is the so-called \textit{Channel-Mixer} (\hyperref[framework]{Fig. 2c}), which operates on each token $f_{j}$ independently and allows communication between different channels and is formulated as:
\begin{equation}
\text{\textit{Channel-Mixer}}(f_j, W^c) = W^c f_j
\label{channel_mixer}
\end{equation}
where $W^c$ denotes the weight matrix with learnable parameters, and $j = 1, 2, \ldots, n$.
Next, the second main sub-module, \textit{Token-Mixer} (\hyperref[framework]{Fig. 2b}), allows communication between different tokens, enabling feature extraction from different spatial locations. Typically, in MLP-based models, the token-mixers formulated as:
\begin{equation}
\text{\textit{Token-Mixer}}(\mathcal{F}, W^t)_j = \sum_{k} W^t_{jk} \odot f_{k},
\label{token_mixer}
\end{equation}
where $W^t$ denotes the corresponding weight matrix for the tokens, and the $\odot$ represents element-wise multiplication. 
Our proposed approach transforms the tokens into wave-like representations to modulate the relationship between tokens and weights dynamically according to their semantic content. In order to represent a token $f_j$ as wave $\tilde{f}_j$ through a wave function, amplitude and phase information are needed:
\begin{equation}
\tilde{f}_j = |f_j| \odot e^{i \theta_j}.
\label{wave_function}
\end{equation}
Here, $i$ denotes the imaginary unit satisfying $i^2 = -1$. The term $|f_j|$ represents the amplitude of the signal. The function $e^{i\theta_j}$ is a periodic function, and $\theta_j$ symbolizes the phase of the signal.
The amplitude $|f_j|$ can be likened to the real-valued feature found in conventional models, with the notable distinction being the application of the absolute value operation.
In the practical implementation, the absolute value operation is omitted and replaced with \ref{channel_mixer} for simplicity.
The phase $\theta_j$ for each token reflects its position within a wave's cycle and can thus be described using fixed parameters, which are learnable during the training phase. Consequently, \ref{channel_mixer} is also utilized for the phase estimation.
Given that \ref{wave_function} characterizes a wave within the complex domain, the Euler formula facilitates embedding tokens within the neural network architecture:
\begin{equation}
\tilde{f}_j = |f_j| \odot \cos \theta_j + i |f_j| \odot \sin \theta_j.
\label{euler}
\end{equation}
Combining \ref{token_mixer} and \ref{euler}, a token is represented as: 
\begin{equation}
f_j = \sum W^t_{jk} f_k \odot \cos \theta_k + W^i_{jk} f_k \odot \sin \theta_k
\label{eq:first_equation}
\end{equation}
\begin{equation}
\Longrightarrow \sum W^t_{jk} f_k \odot \cos(W^c f_k) + W^i_{jk} f_k \odot \sin(W^c f_k)
\label{eq:second_equation}
\end{equation}
where $W^t$, $W^c$ and $W^i$ are learnable weight matrices. The process described, which pertains to wave-like representations, unfolds within the \textit{Token-Mixer}, particularly in the \textit{Wave-Block}. The \textit{Token-Mixer} architecture comprises three blocks: two \textit{Wave-Blocks} and one \textit{Channel-Mixer} operating in parallel. 
The \textit{Vision-MLP} module is structured into four stages. Each stage comprises a sequence consisting of a \textit{Token-Mixer} and a \textit{Channel-Mixer} block, with a normalization layer preceding each. The depth of parallel blocks in each stage is $3$, $4$, $18$, and $3$, respectively. This structure facilitates extracting hierarchical embeddings with corresponding dimensions across stages $64$, $128$, $320$, and $100$.

\subsubsection{Fusion}
For each input frame, the \textit{Vision-MLP} extracts an embedding with a dimensionality of $d=100$. Subsequently, the embeddings derived from the respective frames of a particular video are concatenated to create a unified embedding representation of the original video:
\begin{equation}
\mathcal{V}_D = [d_1 \| d_2 \| \cdots \|d_m], \quad \mathcal{V}_D \in \mathbb{R}^N,
\label{}
\end{equation}
where $m$ denotes the number of frames in a video, and $N$ represents the dimensionality of the final embedding.
Subsequently, the embeddings derived from RGB and synthetic thermal videos are integrated through a weighted fusion process:
\begin{equation}
\mathcal{V}_{Fused} = w_1\cdot\mathcal{V}_{RGB}+w_2\cdot\mathcal{V}_{Thermal},   \quad \mathcal{V}_{Fused}\in\mathbb{R}^N.
\label{eq:fusion}
\end{equation}
The fusion process is founded on combining the corresponding embeddings, utilizing learned weights $w_1$ and $w_2$, which modulate the contributions of the RGB and thermal embeddings, respectively. The weighted addition provides an optimized integration, reflecting the importance of each modality in the final fused representation $\mathcal{V}_{Fused}$.
 
\subsubsection{Transformer}
The fused embeddings are subsequently fed into a transformer-based module comprising self-attention and cross-attention blocks (\hyperref[framework]{Fig. 2e}). The self-attention process is represented as follows:
\begin{equation}
Attention(Q,K,V)= softmax\left( \frac{QK^T}{\sqrt{d_k}}V \right).
\end{equation}
Here, $Q\in\mathbb{R}^{M \times C}$, $K\in\mathbb{R}^{M \times C}$, and $V\in\mathbb{R}^{M \times C}$ represent the Query, Key, and Value matrices, respectively, where  $M$ denotes the input dimension, and $C$ the channel dimension.
Similarly, the cross-attention mechanism employs a dot product operation, but the $Q$ instead of $M \times C$ is $N \times C$, where $N<M$ offers a computational cost reduction.
Each self and cross-attention block incorporates $1$ and $8$ attention heads, respectively, while $4$ parallel blocks comprise the whole \textit{Transformer} module. 
The resulting output embeddings, with a dimensionality of $340$, are employed to complete the final pain assessment through a fully connected neural network.

\subsection{Augmentation Methods}
Two augmentation techniques have been implemented within the framework. First, 
the so-called \textit{Basic} is employed, integrating polarity inversion with noise addition. This method transforms the original input embedding by reversing the polarity of data elements and adding random noise from a Gaussian distribution, creating variability and perturbations. 
Second, the \textit{Masking} involves applying zero-valued masks to the embeddings, nullifying segments of the vectors. The dimensions of the masks are randomly determined, spanning 10\% to 50\% of the embedding's total dimensions, and they are positioned at random locations within the embeddings.

\subsection{Pre-processing}
The pre-processing involved face detection to isolate the facial region. The MTCNN face detector \cite{zhang_2016} was employed, which utilizes multitask cascaded convolutional neural networks for predicting faces and landmarks. It is important to note that the face detector was applied only to the individual RGB frames, and the coordinates of the detected face were applied to the corresponding synthetic thermal frames. The resolution of all frames was set at $224\times 224$ pixels.

\subsection{Pre-training}
For the I2I approach, the \textit{SpeakingFaces} \cite{speakingfaces} dataset was utilized to train the proposed GAN model for translating the RGB to synthetic thermal videos. 
In addition, prior to the automatic pain assessment training process, the \textit{Vision-MLP} and \textit{Transformer} modules were pre-trained. The \textit{Vision-MLP} underwent a three-stage pre-training strategy: initially, it was trained on \textit{DigiFace-1M} \cite{digiface1m} to learn basic facial features. Subsequently, it was trained on \textit{AffectNet} \cite{mollahosseini_hasani_2019} and
\textit{RAF Face Database basic} \cite{li_deng_2017} to learn features related to basic emotions through multi-task learning. Finally, the \textit{Compound Facial Expressions of Emotions Database} \cite{du_tao_2014} and the \textit{RAF Face Database compound} \cite{li_deng_2017} were utilized to learn features of compound emotions in a similar multi-task setting.
The multi-task learning process is described as: 
\begin{equation}
L_{total}= [e^{w1}L_{S_1}+w_{1}]+ [e^{w2}L_{S_2}+w_{2}],
\end{equation}
where $L_S$  is the loss for the corresponding task related to different datasets, and  $w$ represents the learned weights that drive the learning process in minimizing the combined loss $L_{total}$, considering all the individual losses.
The \textit{Transformer} was pre-trained only on the \textit{DigiFace-1M} \cite{digiface1m}, where the input images were flattened into 1D vectors due to its architectural design. 
Table \ref{table:datasets} details the datasets used in the pre-training procedure.

\renewcommand{\arraystretch}{1.2}
\begin{table}
\caption{Datasets utilized for the pretraining process of the framework.}
\label{table:datasets}
\begin{center}
\begin{threeparttable}
\begin{tabular}{ p{3.0cm} p{1.2cm} p{1.2cm} p{1.3cm} }
\toprule
Dataset &\#  samples &\# classes &Task\\
\midrule
\midrule
\textit{SpeakingFaces} \cite{speakingfaces}   &4.58M  &142  &Face$^\varocircle$\\
\textit{DigiFace-1M} \cite{digiface1m}   &1.00M  &10,000  &Face$^\varoast$\\
\textit{AffectNet} \cite{mollahosseini_hasani_2019} &0.40M &8 &Emotion$^\varoast$\\
\textit{Compound FEE-DB}  \cite{du_tao_2014}&6,000 &26 &Emotion$^\varoast$\\
\textit{RAF-DB basic} \cite{li_deng_2017}&15,000 &7 &Emotion$^\varoast$\\
 \textit{RAF-DB compound} \cite{li_deng_2017}&4,000 &11 &Emotion$^\varoast$\\
\bottomrule 
\end{tabular}
\begin{tablenotes}
\scriptsize
\item $\varocircle$: includes face image pairs for the I2I task $\varoast$: includes images for face or emotion recognition tasks
\end{tablenotes}
\end{threeparttable}
\end{center}
\end{table} 


\section{Experimental Evaluation \& Results}
The \textit{BioVid Heat Pain Database} \cite{biovid_2013}, was utilized to evaluate the proposed framework. It comprises facial videos, electrocardiogram, electromyogram, and skin conductance levels from $87$ healthy individuals. The experimental design of the dataset utilized a thermode to induce pain in the participants' right arm, resulting in five distinct intensity levels: no pain (NP), mild pain (P\textsubscript{1}), moderate pain (P\textsubscript{2}), severe pain (P\textsubscript{3}), and very severe pain (P\textsubscript{4}).
Each participant was exposed to each level of pain intensity $20$ times, resulting in $100$ data samples for each modality and $1740$ data samples per class.
We utilized the videos ($5\times 1740 = 8700$) from Part A of \textit{BioVid} in this study.
The pain assessment experiments were structured in binary and multi-level classification settings, evaluating each modality individually and in combination. 
In binary classification, the task was to distinguish between No Pain (NP) and very severe pain (P\textsubscript{4}). In contrast, multi-level classification (MC) involves classifying all pain levels within the dataset.
For evaluation, the leave-one-subject-out (LOSO) cross-validation method was adopted, and performance was assessed based on the accuracy metric.
Table \ref{table:details} presents the framework's training details related to the automatic pain assessment task, and outlines the number of parameters and the computational cost in terms of floating-point operations (FLOPS) for each module.

\renewcommand{\arraystretch}{1.2}
\begin{table}
\centering
\caption{Training details for automatic pain assessment, number of parameters and FLOPS of each module.}
\label{table:details}
\begin{tabular}{ p{2.5cm}  p{2.0cm}  p{2.0cm} }
\toprule
Training Details & Vision-MLP & Transformer \\
\midrule
\midrule
Optimizer: \textit{AdamW}          & Params: 7.35 M       & Params: 7.96 M\\
Learning rate: \textit{2e-5}       & FLOPS: 30.95 G       & FLOPS: 30.90 G\\
LR decay: \textit{cosine}          &                     &\\
Weight decay: 0.1         &                     &\\
Warmup epochs: 5          &                     &\\
Batch size: 32            &                     &\\ \hline
Total            & \multicolumn{2}{l|}{Params: 15.31 Millions \space FLOPS: 61.85 Giga} \\
\bottomrule 
\end{tabular}
\end{table}

\subsection{RGB Videos}
\label{rgb_videos}
In the RGB video modality context, we observed an accuracy of $69.37\%$ for the binary classification task (NP vs. P\textsubscript{4}) and $30.23\%$ for the multi-class classification (MC). Upon intensifying the \textit{Masking} augmentation method to encompass $20-50\%$ of the input embeddings, there was a modest improvement of $0.89\%$ in accuracy for the binary task. In contrast, a decrement was observed in the multi-class task. 
Subsequent extension of training to $300$ epochs, $30-50\%$ for the \textit{Masking} method and $90\%$ probability for both the augmentation methods yielded accuracies of $70.05\%$ and $30.02\%$ for the binary and multi-class tasks, respectively, translating to an average increment marginally below $0.5\%$. Table \ref{table:rgb} presents the classification results. 

\renewcommand{\arraystretch}{1.2}
\begin{table}
\caption{Classification results utilizing the RGB video modality, reported on accuracy \%.}
\label{table:rgb}
\begin{center}
\begin{threeparttable}
\begin{tabular}{ P{1.0cm} P{0.7cm} P{1.0cm}  P{1.0cm}  P{1.3cm}  P{1.0cm}}
\toprule
\multirow{2}[2]{*}{\shortstack{Epochs}}
&\multicolumn{3}{c}{Augmentations} 
&\multicolumn{2}{c}{Task}\\ 
\cmidrule(lr){2-4}\cmidrule(lr){5-6}
&\textit{Basic} &\textit{Masking} &P(Aug) &NP vs P\textsubscript{4} &MC\\
\midrule
\midrule
200  &\checkmark &10-20 &0.7 &69.37 &\textbf{30.23} \\
200  &\checkmark &20-50 &0.7 &\textbf{70.26} &28.50 \\
300  &\checkmark &30-50 &0.9 &70.05 &30.02 \\
\bottomrule 
\end{tabular}
\begin{tablenotes}
\scriptsize
\item Masking: indicates the percentage of the input embedding to which zero-value masking is applied \space\space P(Aug): represents the probability of applying the augmentation methods of Basic \& Masking  \space\space NP: No Pain \space\space P\textsubscript{4}: Very Severe Pain \space\space MC: multiclass pain level
\end{tablenotes}
\end{threeparttable}
\end{center}
\end{table}

\subsection{Synthetic Thermal Videos}
\label{synthetic_thermal_videos}
In the experiments conducted with the synthetic thermal modality under identical experimental conditions, initial accuracies were recorded at $69.97\%$ for the binary task and $30.04\%$ for the multi-class task. An increase in the intensity of the masking method resulted in modest accuracy improvements of $0.23\%$ and $0.46\%$ for the binary and multi-class tasks, respectively. 
Subsequently, final accuracy measurements were $70.69\%$ for the binary task and $29.60\%$ for the multi-class task, culminating in an average increase of $0.28\%$. 
This difference may arise from the challenge of discerning subtle facial changes associated with low-level pain and accompaniment by further corruption from heavier augmentation, which results in diminished performances.
The corresponding results are summarized in Table \ref{table:thermal}. 

\renewcommand{\arraystretch}{1.2}
\begin{table}
\caption{Classification results utilizing the synthetic thermal video modality, reported on accuracy \%.}
\label{table:thermal}
\begin{center}
\begin{threeparttable}
\begin{tabular}{ P{1.0cm} P{0.7cm} P{1.0cm}  P{1.0cm}  P{1.3cm}  P{1.0cm}}
\toprule
\multirow{2}[2]{*}{\shortstack{Epochs}}
&\multicolumn{3}{c}{Augmentations} 
&\multicolumn{2}{c}{Task}\\ 
\cmidrule(lr){2-4}\cmidrule(lr){5-6}
&\textit{Basic} &\textit{Masking} &P(Aug) &NP vs P\textsubscript{4} &MC\\
\midrule
\midrule
200  &\checkmark &10-20 &0.7 &69.97 &30.04 \\
200  &\checkmark &20-50 &0.7 &70.20 &\textbf{30.50} \\
300  &\checkmark &30-50 &0.9 &\textbf{70.69} &29.60 \\
\bottomrule 
\end{tabular}
\begin{tablenotes}
\scriptsize
\item 
\end{tablenotes}
\end{threeparttable}
\end{center}
\end{table}

\subsection{Additional Analysis on RGB \& Synthetic Thermal Videos}
\begin{figure}
\begin{center}
\includegraphics[scale=0.275]{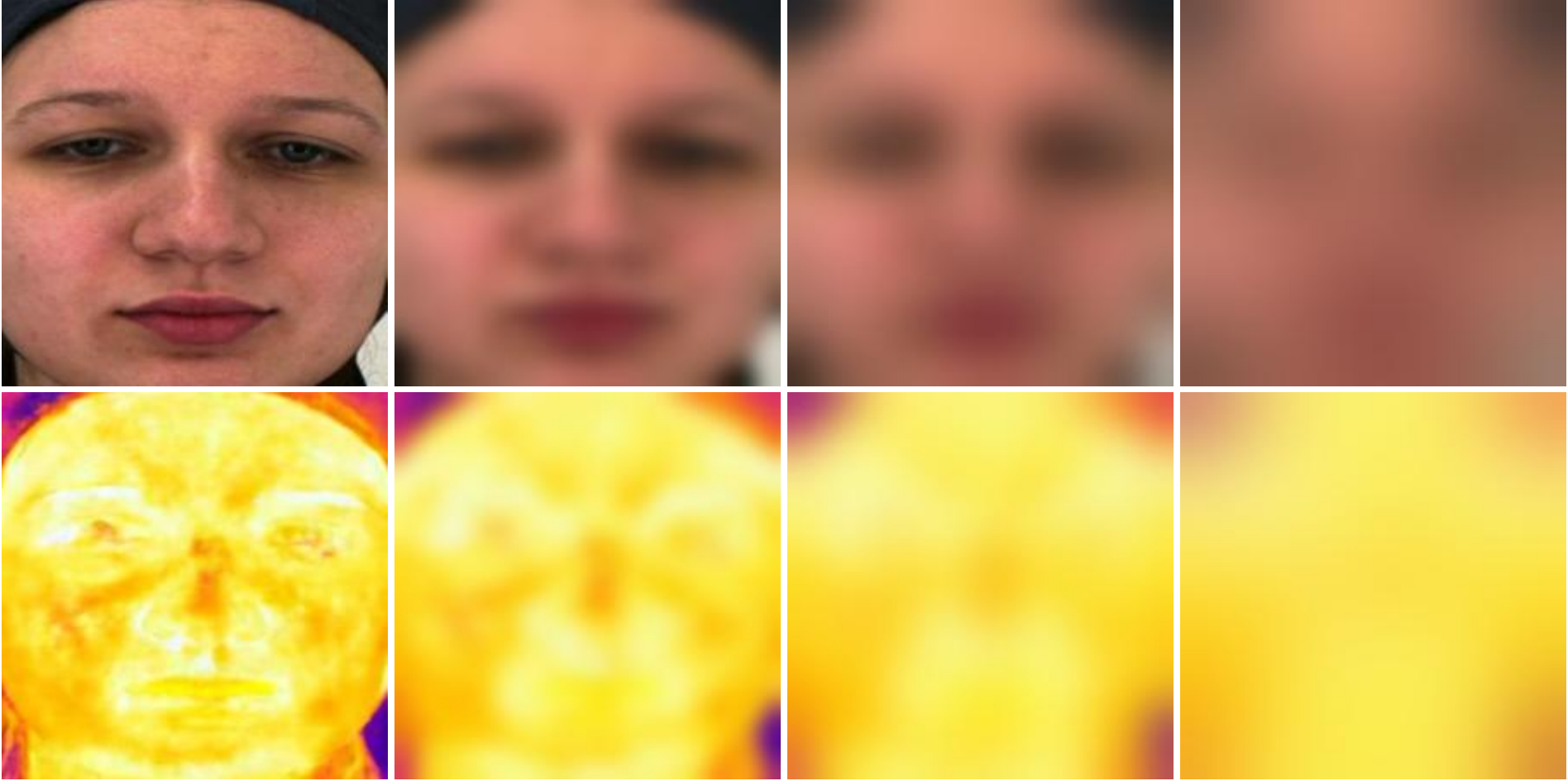} 
\end{center}
\caption{Progressive blurring of RGB and synthetic thermal facial imagery: a sequence illustrating varying degrees of Gaussian blur applied, with kernel sizes incrementally adjusted from $k = 0$ (clear) to $k = 191$ (heavily blurred).}
\label{faces}
\end{figure}

The findings from \ref{rgb_videos} and \ref{synthetic_thermal_videos} revealed a notable circumstance where the performance metrics for the RGB and synthetic thermal modalities are remarkably similar. Specifically, the highest recorded accuracies for the RGB modality were $70.26\%$ and $30.23\%$ for the NP vs. P4 and MC tasks, respectively. Correspondingly, the peak accuracies for the synthetic thermal modality were $70.69\%$ and $30.50\%$. On average, the performances from the thermal videos are approximately $1\%$ superior to those of the RGB modality.
This outcome was unexpected, given that the synthetic modality was initially presumed to be less effective than the original. 
This prompted an exploration into the reason synthetic modalities exhibit comparable or superior performance to the original RGB modality.
A primary question was regarding the richness and effectiveness of the thermal-related information incorporated in the synthetic videos. 
The hypothesis suggested that reducing facial expressions in the thermal videos could allow a more explicit assessment of the thermal information.
Gaussian blurring was progressively applied to RGB and synthetic thermal videos (refer to Fig. \ref{faces}), with kernel sizes $k$ incrementally adjusted from $0$ to $191$. Similar, albeit less time-intensive, experiments to \ref{rgb_videos}, \ref{synthetic_thermal_videos} were conducted.

\begin{figure}
\begin{center}
\includegraphics[scale=0.12]{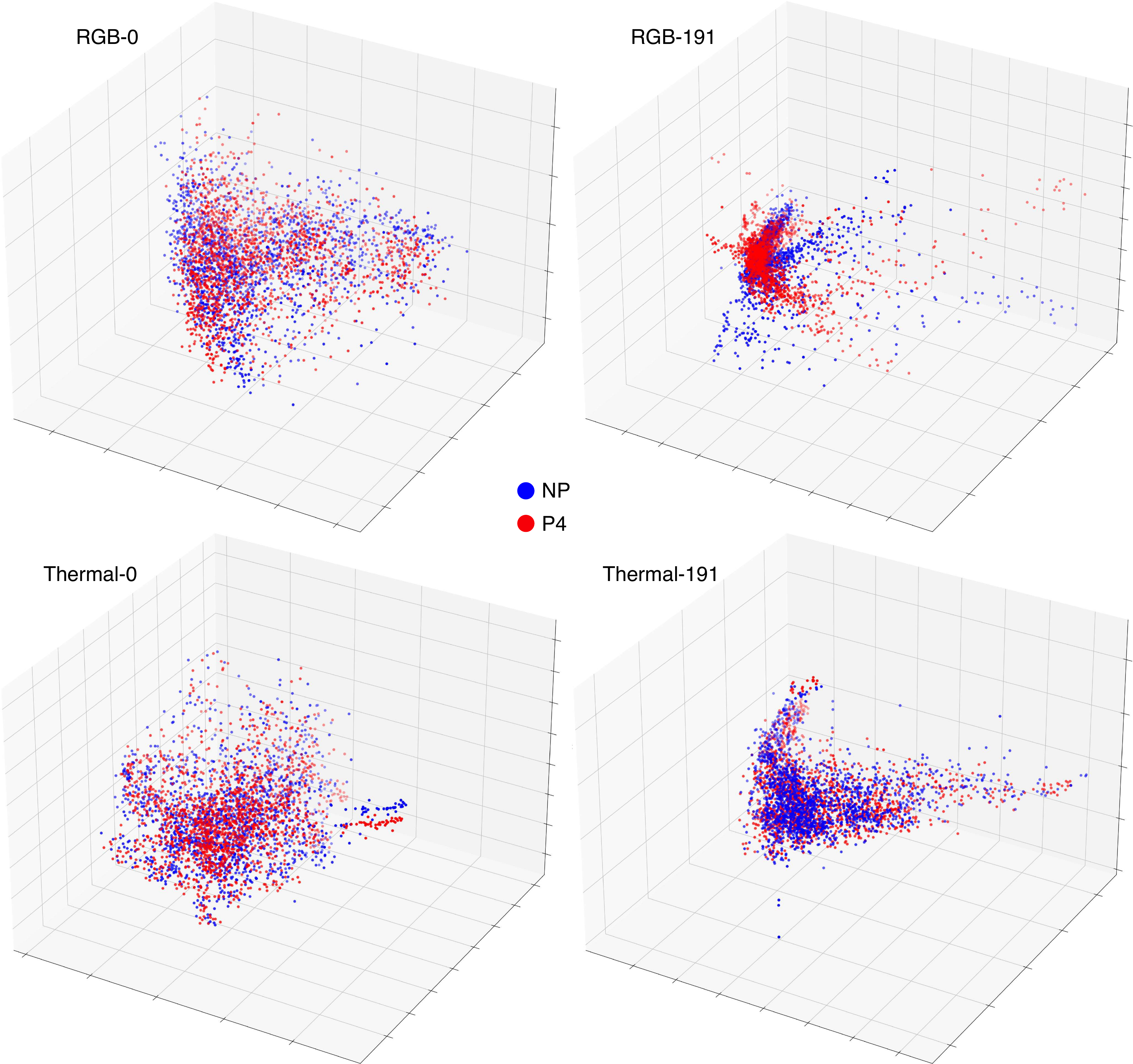} 
\end{center}
\caption{3D embedding space distributions of NP (no pain) and P4 (very severe pain) classes in RGB and synthetic thermal videos, for $k = 0$ (clear) and $k = 191$ (heavily blurred).}
\label{emb}
\end{figure}

Table \ref{table:blur} shows that with a kernel size of $k=0$, the performance disparity of $0.47\%$ (favoring the thermal modality) aligns with prior experimental outcomes. As blurring intensifies to $k=41$, this discrepancy marginally increases to $0.49\%$. Notably, at $k=91$, the divergence expands to $2.13\%$ and intensifies to $5.90\%$ when the blur peaks at $k=191$ (heavily blurred).
The classification performances demonstrated that by diminishing the visibility of facial expressions through blurring, the synthetic thermal videos resulted in superior performance compared to the RGB, with figures of $66.24\%$ over $60.34\%$. Additionally, as the kernel size increased from $k=0$ to $k=191$, the decline in accuracy rates for the synthetic thermal and RGB modalities was $1.81\%$ and $7.13\%$, respectively. This suggests that the residual information in the synthetic modality, essentially the visually represented facial temperature, remains intact or minimally influenced.
Fig. \ref{emb} depicts the embedding distribution for the RGB and synthetic thermal modality for $k=0$ and $k=191$.
Although the separation of the data points is not clear, we observe a distinct difference in the distribution. 
For $k=191$, the RGB embeddings are centralized and probably overlap, and a plethora of points are notably spread away from the central mass without a clear pattern. Respectively, the data points are much more uniformly spreading for the synthetic modality, suggesting potentially better differentiation between classes.

\renewcommand{\arraystretch}{1.2}
\begin{table}
\caption{Classification results utilizing the RGB  \& the synthetic thermal video modality  reported, on accuracy \%.}
\label{table:blur}
\begin{center}
\begin{threeparttable}
\begin{tabular}{ P{0.7cm} P{0.8cm} P{0.7cm} P{0.5cm} P{0.7cm}  P{1.0cm}  P{1.1cm} }
\toprule
\multirow{2}[2]{*}{\shortstack{Epochs}}
&\multirow{2}[2]{*}{\shortstack{Modality}}
&\multirow{2}[2]{*}{\shortstack{Blur}}
&\multicolumn{3}{c}{Augmentations} 
&\multicolumn{1}{c}{Task}\\ 
\cmidrule(lr){4-6}\cmidrule(lr){7-7}
& & &\textit{Basic} &\textit{Masking} &P(Aug) &NP vs P\textsubscript{4}\\
\midrule
\midrule
100  &RGB     &0 &\checkmark &10-20 &0.7 &67.47 \\
100  &Thermal &0 &\checkmark &10-20 &0.7 &68.05  \\\hline
100  &RGB     &41 &\checkmark &10-20 &0.7 &66.61 \\
100  &Thermal &41 &\checkmark &10-20 &0.7 &67.10  \\\hline
100  &RGB     &91 &\checkmark &10-20 &0.7 &64.80 \\
100  &Thermal &91 &\checkmark &10-20 &0.7 &66.93  \\\hline
100  &RGB &191 &\checkmark &10-20 &0.7 &60.34 \\
100  &Thermal &191 &\checkmark &10-20 &0.7 &66.24 \\
\bottomrule 
\end{tabular}
\begin{tablenotes}
\scriptsize
\item Blur: Gaussian blurring with kernel sizes $k$
\end{tablenotes}
\end{threeparttable}
\end{center}
\end{table}

\subsection{Fusion}
Three fusion methods were assessed in the context of multimodal analysis for RGB and synthetic thermal videos. The approach outlined in \ref{eq:fusion} was initially applied, utilizing learned weights $w_1$ and $w_2$ to scale the respective modalities. Additionally, a second method was employed where a third weight, $w_3$, was introduced, resulting in $w_3\cdot(w_1\cdot\mathcal{V}_{RGB}+w_2\cdot\mathcal{V}_{Thermal})$. Lastly, a method without learned weights was explored, directly adding the embedding vectors from both modalities. Table \ref{table:fusion} presents the corresponding results. The absence of weights resulted in $64.92\%$ and $26.40\%$ accuracy for the binary and multi-class tasks, respectively. The integration of the three weights resulted in a decrease of $0.5\%$ in accuracy for both tasks, whereas the application of weights $w_1$ and $w_2$ yielded the highest performance, with accuracies reaching $65.08\%$ and $26.50\%$ for the binary and multi-class tasks, respectively.
By applying weights $w_1$ and $w_2$ and increasing the training period from $100$ to $300$ epochs while maintaining consistent augmentation settings, accuracies of $69.50\%$ and $29.80\%$ were achieved for the binary and multi-class tasks, respectively.
Further extending the training time to $500$ epochs without encountering any overfitting phenomena improved performance, reaching accuracies of $71.03\%$ and $30.70\%$ for the respective tasks.

\renewcommand{\arraystretch}{1.2}
\begin{table}
\caption{Classification results utilizing the fusion of RGB \& synthetic thermal video modality, reported on accuracy \%.}
\label{table:fusion}
\begin{center}
\begin{threeparttable}
\begin{tabular}{ P{0.7cm} P{1.0cm} P{0.5cm} P{0.7cm}  P{1.0cm}  P{1.1cm}  P{0.7cm}}
\toprule
\multirow{2}[2]{*}{\shortstack{Epochs}}
&\multirow{2}[2]{*}{\shortstack{Fusion\\weights}}
&\multicolumn{3}{c}{Augmentations} 
&\multicolumn{2}{c}{Task}\\ 
\cmidrule(lr){3-5}\cmidrule(lr){6-7}
& &\textit{Basic} &\textit{Masking} &P(Aug) &NP vs P\textsubscript{4} &MC\\
\midrule
\midrule
100  &-- &\checkmark &10-20 &0.7 &64.92 &26.40 \\
100  &W2 &\checkmark &10-20 &0.7 &65.08 &26.50 \\
100  &W3 &\checkmark &10-20 &0.7 &64.42 &25.90 \\
\hline
300  &W2 &\checkmark &10-20 &0.7 &69.50 &29.80 \\
500  &W2 &\checkmark &10-20 &0.7 &\textbf{71.03} &\textbf{30.70} \\
\bottomrule 
\end{tabular}
\begin{tablenotes}
\scriptsize
\item W2: utilization of [w\textsubscript{1},w\textsubscript{2}] \space\space 
W3: utilization of [w\textsubscript{1},w\textsubscript{2},w\textsubscript{3}]
\end{tablenotes}
\end{threeparttable}
\end{center}
\end{table}

\section{Comparison with Existing Methods}
This section compares the proposed method with other existing approaches in the literature. The evaluation utilizes Part A of the \textit{BioVid} dataset, involving all $87$ subjects, and follows the same validation protocol, LOSO cross-validation. Table \ref{table:sota} presents the corresponding results. 
The proposed vision-based method, utilizing RGB and synthetic thermal modalities, demonstrated performances comparable to or exceeding that of previous methods.
Compared to the findings reported in studies \cite{werner_hamadi_walter_2017,zhi_wan_2019,thiam_kestler_schenker_2020,tavakolian_bordallo_liu_2020}, improved accuracy was attained in binary and multi-level tasks.
It is noted that the authors in \cite{werner_2016} reported accuracies of $72.40\%$ and $30.80\%$, showing an improvement of $1.37\%$ and $0.10\%$ over our results. In the study \cite{gkikas_tsiknakis_embc}, the authors achieved the highest reported results, employing a transformer-based architecture.

Furthermore, in Table \ref{table:mintpain}, we compare our findings with those from the study \cite{haque_2018}, in which the authors introduce the \textit{MIntPAIN} dataset, including both RGB and thermal videos for automatic pain assessment across five intensity levels.
We observe that the accuracy of the RGB and thermal modalities is particularly similar, at $18.55\%$ and $18.33\%$, respectively. This outcome mirrors our findings, where performance between the two modalities—RGB and synthetic thermal was similarly aligned.
By fusing the modalities, the authors achieved an increase of $30.77\%$, marking an improvement of over $12\%$. This contrasts with our results, where the increase was marginal.
However, it is important to note that the performances of the unimodal approaches in  \cite{haque_2018} were below the guess prediction threshold of $20\%$, and only through the fusion of these modalities did the performance surpass it.

\renewcommand{\arraystretch}{1.2}
\begin{table}
\caption{Comparison of studies that utilized \textit{BioVid}, videos, and LOSO cross-validation, reported on accuracy\%.}
\label{table:sota}
\begin{center}
\begin{threeparttable}
\begin{tabular}{ P{2.5cm} P{2.5cm} P{1.1cm}  P{0.7cm}}
\toprule
\multirow{2}[2]{*}{\shortstack{Study}}
&\multirow{2}[2]{*}{\shortstack{Method}}
&\multicolumn{2}{c}{Task}\\ 
\cmidrule(lr){3-4}
& &NP vs P\textsubscript{4} &MC\\
\midrule
\midrule
Werner \textit{et al.}\cite{werner_2016} &Deep RF &72.40&30.80\\
Werner \textit{et al.}\cite{werner_hamadi_walter_2017} &RF &70.20 &--\\
Zhi \textit{et al.} \cite{zhi_wan_2019} &SLSTM &61.70 &29.70\\
Thiam \textit{et al.} \cite{thiam_kestler_schenker_2020} &2D CNN, biLSTM &69.25 &--\\
Tavakolian \textit{et al.} \cite{tavakolian_bordallo_liu_2020} &2D CNN &71.00 &--\\
Gkikas \textit{et al.}\cite{gkikas_tsiknakis_embc} &Vision-Transformer &73.28 &31.52\\
Our &Vision-MLP &71.03 &30.70\\
\bottomrule 
\end{tabular}
\end{threeparttable}
\end{center}
\end{table}

\renewcommand{\arraystretch}{1.1}
\begin{table}
\caption{Comparison with the \textit{MIntPAIN} dataset, reported on accuracy\%.}
\label{table:mintpain}
\begin{center}
\begin{threeparttable}
\begin{tabular}{ P{2.0cm} P{1.5cm} P{1.5cm} P{1.0cm}}
\toprule
\multirow{2}[2]{*}{\shortstack{Study}}
&\multirow{2}[2]{*}{\shortstack{Dataset}}
&\multirow{2}[2]{*}{\shortstack{Modality}}
&\multicolumn{1}{c}{Task}\\ 
\cmidrule(lr){4-4}
& & &MC\\
\midrule
\midrule
\multirow{3}{*}{Haque \textit{et al.} \cite{haque_2018}} &\multirow{3}{*}{MIntPAIN} &RGB       &18.55\\
                                                                   & &Thermal$^{\circ}$ &18.33\\
                                                                   & &Fusion  &30.77 \\
\hline
\multirow{3}{*}{Our} &\multirow{3}{*}{BioVid} &RGB       &30.02\\
                             & &Thermal$^{\star}$ &29.69\\
                             & &Fusion  &30.70\\              
\bottomrule 
\end{tabular}
\begin{tablenotes}
\scriptsize
\item $\circ$:real\space $\star$: synthetic
\end{tablenotes}
\end{threeparttable}
\end{center}
\end{table}

\section{Conclusion}
This study explored the generation of synthetic thermal imagery from GAN models to evaluate its effectiveness in the context of automatic pain assessment. Furthermore, a novel framework based on \textit{Vision-MLP} was introduced, complimented by a \textit{Transformer} module serving as the core of the assessment system.
The conducted experiments underscored the efficacy of the synthetic thermal modality, showcasing performances comparable to or surpassing those of the original RGB modality. Moreover, this study examined the underlying factors contributing to this effectiveness, particularly focusing on the role of temperature color representations.
Additionally, the integration of the two vision modalities was analyzed using various fusion techniques. 
It should be emphasized that further optimization and experimentation, particularly with the multimodal approach, have the capacity to yield enhanced results.
We believe that the generation and integration of synthetic modalities, such as thermal imagery, in an automatic pain assessment framework holds significant potential, and additional exploration and research are needed.

\section*{Ethical Impact Statement}
This research employed the \textit{BioVid Heat Pain Database} \cite{biovid_2013} to evaluate the proposed methods. The data were recorded according to the ethical guidelines of Helsinki (\textit{ethics committee: 196/10-UBB/bal}).
Prior to commencing data collection, each participant's pain threshold (where sensation transitions from heat to pain) and tolerance threshold (the moment when pain becomes unbearable) were determined.
The facial images presented in this study are from participants who have consented to their use for illustrative purposes within a scientific research context.
This study aims to introduce a pain assessment framework designed to facilitate continuous patient monitoring while reducing human biases.
However, it is essential to recognize that real-world applications, especially in clinical settings, might present challenges, necessitating further experimentation and comprehensive evaluation through clinical trials before deployment.

In addition, this study utilized the \textit{SpeakingFaces} \cite{speakingfaces} dataset for the image-to-image translation process. The data was collected according to the ethical guidelines of the Declaration of Helsinki,
and with the approval from the Institutional Research Ethics Committee at Nazarbayev University. All participants were volunteers who were fully informed about the data collection procedures and the intended use of identifiable images, which will be distributed as part of a dataset. Each participant provided their permission by signing informed consent forms.

Furthermore, several datasets were utilized to pretrain the proposed pain assessment framework. 
The \textit{DigiFace-1M} \cite{digiface1m} is a synthetic dataset where $511$ initial face scans were obtained with consent and employed to build a parametric face geometry and texture library model. All the identities and samples were generated from these source data.   
The \textit{AffectNet} \cite{mollahosseini_hasani_2019} dataset is compiled using search engine queries. The original paper does not explicitly detail ethical compliance measures such as adherence to the Declaration of Helsinki or informed consent procedures.
The original paper of \textit{Compound FEE-DB}  \cite{du_tao_2014} does not mention ethical compliance measures, but only that the subjects were recruited from the Ohio State University area and received a monetary reward for participating.
The \textit{RAF-DB} \cite{li_deng_2017} dataset was compiled using the Flickr image hosting service. Although Flickr hosts both public and privately shared images, the authors do not explicitly mention the type of the downloaded images. 

\section*{Acknowledgement}
Research supported by the \textit{ODIN} project that has received funding from
the European Union’s Horizon 2020 research and innovation program under
grant agreement No $101017331$.

\bibliographystyle{IEEEtran}
\bibliography{library}

\appendix

\subsection*{Supplementary Metrics}

\renewcommand{\arraystretch}{1.2}
\begin{table}[h!]
\caption{Classification results utilizing the RGB video modality, reported on recall and F1 score.}
\label{table:appendix_rgb}
\begin{center}
\begin{threeparttable}
\begin{tabular}{ P{0.7cm} P{0.4cm} P{0.7cm}  P{0.7cm} P{0.8cm} P{1.1cm} P{0.7cm}}
\toprule
\multirow{2}[2]{*}{\shortstack{Epochs}}
&\multicolumn{3}{c}{Augmentations} 
&\multirow{2}[2]{*}{\shortstack{Metric}}
&\multicolumn{2}{c}{Task}\\ 
\cmidrule(lr){2-4}\cmidrule(lr){6-7}
&Basic &Masking &P(Aug) & &NP vs P\textsubscript{4} &MC\\
\midrule
\midrule
\multirow{2}{*}{200}    &\multirow{2}{*}{\checkmark} 
&\multirow{2}{*}{30-50} &\multirow{2}{*}{0.9} &Recall    &71.29  &29.61\\
                                          && &  &F1      &68.53  &27.22\\                     
\hline
\multirow{2}{*}{200}    &\multirow{2}{*}{\checkmark} 
&\multirow{2}{*}{30-50} &\multirow{2}{*}{0.9} &Recall    &71.93  &24.43\\
                                              && & &F1   &69.61  &23.78\\
\hline
\multirow{2}{*}{300}    &\multirow{2}{*}{\checkmark} 
&\multirow{2}{*}{30-50} &\multirow{2}{*}{0.9} &Recall    &71.34  &30.64\\
                                          && &  &F1      &69.65  &26.12\\                      
\bottomrule 
\end{tabular}
\begin{tablenotes}
\scriptsize
\item  \space 
\end{tablenotes}
\end{threeparttable}
\end{center}
\end{table}

\renewcommand{\arraystretch}{1.2}
\begin{table}[ht]
\caption{Classification results utilizing the synthetic thermal video modality, reported on recall and F1 score.}
\label{table:appendix_thermal}
\begin{center}
\begin{threeparttable}
\begin{tabular}{ P{0.7cm} P{0.4cm} P{0.7cm}  P{0.7cm} P{0.8cm} P{1.1cm} P{0.7cm}}
\toprule
\multirow{2}[2]{*}{\shortstack{Epochs}}
&\multicolumn{3}{c}{Augmentations} 
&\multirow{2}[2]{*}{\shortstack{Metric}}
&\multicolumn{2}{c}{Task}\\ 
\cmidrule(lr){2-4}\cmidrule(lr){6-7}
&Basic &Masking &P(Aug) & &NP vs P\textsubscript{4} &MC\\
\midrule
\midrule
\multirow{2}{*}{200}    &\multirow{2}{*}{\checkmark} 
&\multirow{2}{*}{30-50} &\multirow{2}{*}{0.9} &Recall    &72.04  &28.80\\
                                          && &  &F1      &69.16  &26.45\\                     
\hline
\multirow{2}{*}{200}    &\multirow{2}{*}{\checkmark} 
&\multirow{2}{*}{30-50} &\multirow{2}{*}{0.9} &Recall    &72.18  &30.89\\
                                              && & &F1   &69.44  &26.45\\
\hline
\multirow{2}{*}{300}    &\multirow{2}{*}{\checkmark} 
&\multirow{2}{*}{30-50} &\multirow{2}{*}{0.9} &Recall    &72.52  &24.96\\
                                              && & &F1   &70.01  &23.43\\                  
\bottomrule 
\end{tabular}
\begin{tablenotes}
\scriptsize
\item  \space 
\end{tablenotes}
\end{threeparttable}
\end{center}
\end{table}

\renewcommand{\arraystretch}{1.2}
\begin{table}[ht]
\caption{Classification results utilizing the fusion of RGB \& synthetic thermal video modality, reported on recall and F1 score.}
\label{table:appendix}
\begin{center}
\begin{threeparttable}
\begin{tabular}{ P{0.6cm} P{0.8cm} P{0.33cm} P{0.68cm}  P{0.63cm} P{0.8cm} P{1.05cm}  P{0.5cm}}
\toprule
\multirow{2}[2]{*}{\shortstack{Epochs}}
&\multirow{2}[2]{*}{\shortstack{Fusion\\weights}}
&\multicolumn{3}{c}{Augmentations} 
&\multirow{2}[2]{*}{\shortstack{Metric}}
&\multicolumn{2}{c}{Task}\\ 
\cmidrule(lr){3-5}\cmidrule(lr){7-8}
& &Basic &Masking &P(Aug) & &NP vs P\textsubscript{4} &MC\\
\midrule
\midrule
\multirow{2}{*}{100}    &\multirow{2}{*}{--} &\multirow{2}{*}{\checkmark} 
&\multirow{2}{*}{30-50} &\multirow{2}{*}{0.9} &Recall    &67.05  &21.68\\
                                          &&& &  &F1     &62.96  &18.29\\  
\hline                               
\multirow{2}{*}{100}    &\multirow{2}{*}{W2} &\multirow{2}{*}{\checkmark} 
&\multirow{2}{*}{30-50} &\multirow{2}{*}{0.9} &Recall    &68.72  &21.69\\
                                          &&& &  &F1     &62.98  &19.35\\ 
\hline         
\multirow{2}{*}{100}    &\multirow{2}{*}{W3} &\multirow{2}{*}{\checkmark} 
&\multirow{2}{*}{30-50} &\multirow{2}{*}{0.9} &Recall    &66.12  &23.12\\
                                          &&& &  &F1     &59.72  &19.67\\                     
\hline
\multirow{2}{*}{300}    &\multirow{2}{*}{W2} &\multirow{2}{*}{\checkmark} 
&\multirow{2}{*}{30-50} &\multirow{2}{*}{0.9} &Recall    &71.40  &26.39\\
                                              &&& & &F1  &68.82  &26.18\\
\hline
\multirow{2}{*}{500}    &\multirow{2}{*}{W2} &\multirow{2}{*}{\checkmark} 
&\multirow{2}{*}{10-20} &\multirow{2}{*}{0.7} &Recall    &73.20  &29.69\\
                                            & & & & &F1  &70.30  &27.84\\                    
\bottomrule 
\end{tabular}
\begin{tablenotes}
\scriptsize
\item  \space 
\end{tablenotes}
\end{threeparttable}
\end{center}
\end{table}

\end{document}